\documentclass[sigconf,nonacm]{acmart}

\usepackage{subcaption}
\usepackage{makecell}
\usepackage{algorithm}
\usepackage{algpseudocode}
\usepackage{multirow}
\usepackage{graphicx}
\usepackage{subcaption}
\usepackage{ragged2e}
\usepackage{bm}
\usepackage{enumitem}
\usepackage{amsmath}

\usepackage[skip=3pt]{caption} % default 10pt
\newtheorem{definition}{Definition}

\AtBeginDocument{%
  }

\setcopyright{acmlicensed}
\copyrightyear{2018}
\acmYear{2018}
\acmDOI{XXXXXXX.XXXXXXX}
\acmConference[Conference acronym 'XX]{Make sure to enter the correct
  conference title from your rights confirmation email}{June 03--05,
  2018}{Woodstock, NY}

\begin{document}

\title{SCAR: A Characterization Scheme for Multi-Modal Dataset}

\author{Ri Su$^{1}$,
        Zhao Chen$^{1,*}$,
        Caleb Chen Cao$^{2,*}$,
        Nan Tang$^{1}$,
        Lei Chen$^{1,2}$}
\affiliation{
\institution{
{$^{1}$} HKUST (GZ),
{$^{2}$}The Hong Kong University of Science and Technology\\
rsu704@connect.hkust-gz.edu.cn, \{chenzhao, nantang\}@hkust-gz.edu.cn, cao@ust.hk, leichen@cse.ust.hk}
}
\thanks{$^{*}$Co-Corresponding author.}

\begin{abstract}
      Foundation models exhibit remarkable generalization across diverse tasks, largely driven by the characteristics of their training data. 
Recent data-centric methods like pruning and compression aim to optimize training but offer limited theoretical insight into how data properties affect generalization, especially the data characteristic in sample scaling.
Traditional perspectives further constrain progress by focusing predominantly on data quantity and training efficiency, often overlooking structural aspects of data quality.
In this study, we introduce \textbf{SCAR}, a principled scheme for characterizing the intrinsic structural properties of datasets across four key measures: Scale, Coverage, Authenticity, and Richness. 
Unlike prior data-centric measures, SCAR captures stable characteristics that remain invariant under dataset scaling, providing a robust and general foundation for data understanding. 
Leveraging these structural properties, we introduce \textbf{Foundation Data}—a minimal subset that preserves the generalization behavior of the full dataset without requiring model-specific retraining. 
We model single-modality tasks as step functions and estimate the foundation data size distribution to capture step-wise generalization bias across modalities in the target multi-modal dataset.
Finally, we develop a SCAR-guided data completion strategy based on this generalization bias, which enables efficient, modality-aware expansion of modality-specific characteristics in multimodal datasets.
Experiments across diverse multi-modal datasets and model architectures validate the effectiveness of SCAR in predicting data utility and guiding data acquisition.
Code is available at \url{https://github.com/McAloma/SCAR}.

\end{abstract}


\maketitle

% \newpage
\section{Introduction}
Foundation Models (FMs) \cite{foundation, gpt} have emerged as a unifying paradigm~\cite{dino, clip}, offering transferable representations that power several downstream tasks in modern data-centric systems \cite{orr2022data}.
Unlike traditional machine learning pipelines reliant on clean, structured datasets, FMs utilize vast and heterogeneous corpora, enabling broad task coverage \cite{madden2024databases} with minimal task-specific supervision. 
However, this reliance on large-scale, heterogeneous, and weakly curated data \cite{datagat} renders FMs’ performance critically dependent on—and highly sensitive to—training data characteristics \cite{glavic2024towards}. 
In practice, large-scale datasets often contain substantial noise, bias, and distributional shifts. Under these conditions, simply increasing data volume proves insufficient for enhancing robustness or generalization \cite{uselisdoes}.

Recent data-centric methods \cite{zha2023data} enhance model performance by improving data quality rather than changing model scale or architecture.
A key research direction focuses on label noise by modeling annotation reliability and training dynamics. Confidence-based approaches, such as Confident Learning \cite{northcutt2021confident}, estimate label correctness to filter noisy samples. Extensions like confidence regularization \cite{cheng2021learning} and noise transition modeling \cite{jiang2021information} incorporate uncertainty into training. However, they rely on model predictions, which may be unreliable under noise or imbalance, leading to biased filtering.
Meanwhile, data distillation methods \cite{wang2018dataset} improve data efficiency by selecting informative examples. Techniques such as data quantization \cite{zhao2024dataset, zhou2023dataset} reduce redundancy and distribution bias, while others use optimization objectives to build representative subsets. Yet, many are model- or task-specific and offer limited theoretical understanding of how intrinsic structural properties, such as the dispersion of data across task subspaces or the concentration of samples in feature space, affect the generalization or scaling behavior of FMs.

\begin{figure}[t]
    \centering
    \includegraphics[width=0.475\textwidth]{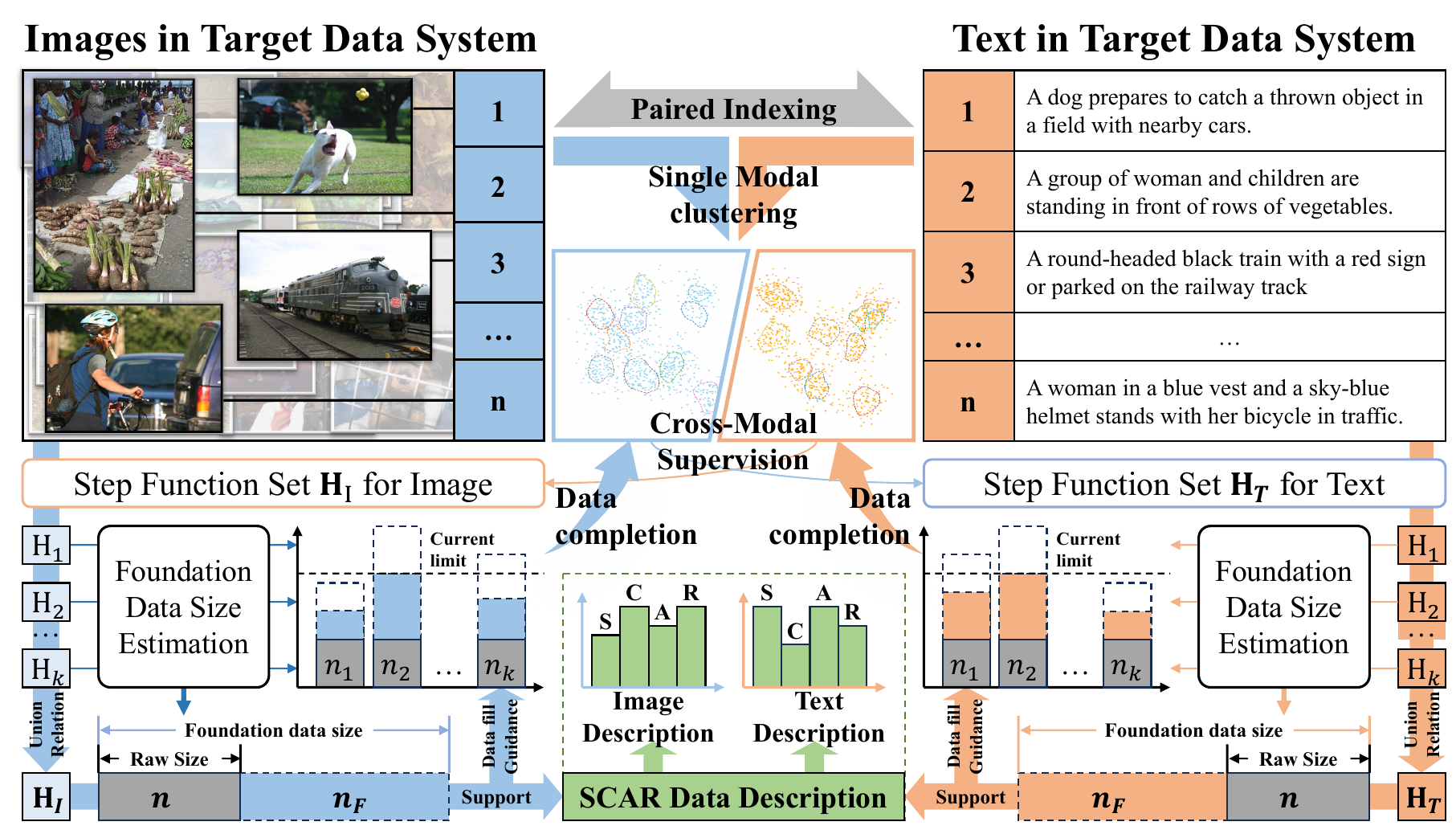}
    \caption{An illustration of the SCAR data characterization scheme based on an example of an image-text dataset system. This scheme quantifies dataset utility through four measures: Scale, Coverage, Authenticity, and Richness, and leverages the concept of Foundation Data for theoretical estimation and guides efficient data description and completion.}
    \Description{1234}
    \label{fig:structure}
\end{figure}

From a theoretical standpoint, the Probably Approximately Correct (PAC) learning \cite{zhang2023mathematical} provides generalization guarantees for risk-based learning, typically through concentration inequalities and union bounds. 
Such theoretical guarantees offer sample complexity bounds that are agnostic to model architecture and optimization procedures. 
However, classical assumptions often break down in real-world data, which are structured, sparse, and inputs may correspond to multiple valid outputs \cite{DBLP:conf/iclr/ZhangBHRV17}. 
In such contexts, the generalization is not only determined by model capacity but also influenced by the characteristics of the training data \cite{nagarajan2019uniform}.

In this study, we revisit learning theory with data-centric practice and further characterize the performance of each modality in a dataset via cross-modal signal supervision. 
Firstly, we theoretically define Foundation Data Size (FDS) as the minimal data scale required to ensure generalization under a distributional perspective. 
Unlike data distillation, which extracts a core subset from existing data based on model-specific signals, FDS estimates the minimum dataset size needed at a macro level, independent of training dynamics. 
This provides a model-agnostic view into the intrinsic scalability and utility of a dataset across scaling levels.
Secondly, we propose SCAR, a data quality scheme with four measures—Scale, Coverage, Authenticity, and Richness—to characterize data utility in a single step-function and general perspective. 
Then, we design a principled data completion framework guided by FDS estimation, where SCAR serves as a global diagnostic of data quality and informs modality-specific augmentation strategies. 
Finally, we extend the framework to multi-modal data by using cluster-derived labels from one modality to supervise another, enabling effective evaluation of various datasets.
Fig.\ref{fig:structure} illustrates the process of SCAR and its relationship to multi-modal data characteristics enhancement.
The key contributions are as follows:
\begin{itemize}[leftmargin=*]
    \item We derive a theoretical framework for estimating the size of foundation data with SCAR measures, which describes how performance data characteristics change with expansion.
    \item The SCAR data quality description index was proposed, which quantifies data quality in terms of Scale, Coverage, Authenticity, and Richness.
    \item By estimating the size of foundation data, we provide a data completion method, which can guide the data collection process more efficiently and effectively.
\end{itemize}

Through extensive experiments across various tasks, datasets, and models, we demonstrate that the SCAR index effectively captures data quality in terms of its impact on task performance on different data formats and that the foundation data size estimation provides a practical and theoretical basis for data completion.

% \newpage
\section{Preliminaries}
Consider an unknown data distribution $\mathcal{D}$, and an ideal encoding function $\phi^*(\cdot)$ (e.g., a semantic representation or compression function in related literature).
We assume access to an oracle $\mathcal{O}_m$ for modality $m$, which, upon each call, returns a data point $X \sim \mathcal{D}$ together with its latent knowledge representation $ K=\phi^*(X) \in\mathbb {R}^d$.
This representation $K$ is assumed to be lossless, which means that there exists an inverse mapping $\phi^{*-1}$ such that $\phi^{*-1}(K) = X \sim \mathcal{D}$. 

Given this representation, the binary label $Y$ is then defined as a threshold linear projection of $K$, specifically via a Heaviside step function $H(\cdot)$ applied to a fixed linear transformation of $K$:
\begin{equation*}
    Y = H(WK + B) = 
    \begin{cases}
    0, & WK + B < 0, \\
    1, & WK + B \geq 0,
    \end{cases}
\end{equation*}
where $W \in \mathbb{R}^{d}$ and $B \in \mathbb{R}$. This formulation models threshold-based decisions between samples and matches the interpretable structure often needed in semantic tasks. 
In a general perspective, a specific task $\mathbf{H} = \{H_j\}$ consists of $k$ such step functions.
The concept space $\mathcal{H}$ is then defined as the set of all such compositions:
\begin{equation*}
    \mathcal{H} = \left\{ \mathbf{H}(K)\;\middle|\; \mathbf{H}(K) = \{H_j(W_jK + B_j)\},\; j=1,\dots,k \right\}.
\end{equation*}
The dataset used for learning is generated by repeatedly calling $\mathcal{O}$ to obtain samples $\{X_i, K_i\}_{i=1}^n$.
For a given step function set $\mathbf{H}^* = \{H^*_j\}$, the true binary label and resulting dataset are:
\begin{equation*}
    Y_{ij} = H^*_j(W^*K_i + B^*), \;\;\; \mathcal{S}_n = \left\{X_i, \{Y_{ij}\}_{j=1}^k \right\}_{i=1}^n,
\end{equation*}
where $k$ is the number of target step functions, and $n$ is the number of samples in the dataset. 
The goal of the ERM-based learning algorithm is to find a function $f(\cdot) \approx \mathbf{H}^*(\phi^*(\cdot))$ that approximates the minimum empirical error $\hat{err}_{\mathcal{S}_n}(f(\cdot))$.
When given a $\phi(\cdot)$ the function class $\{f(\cdot)\}$ corresponds to the $\mathcal{H}$, and its size is $|\mathcal{H}|$.

To analyze the general error of the signal function set $\mathbf{H}$, let $E$ denote the event $err_\mathcal{D}(\mathbf{H}) \geq \epsilon$, and let $E_j$ denote $err_\mathcal{D}(H_j) \geq \epsilon$. 
Since the same dataset is shared among all $H_j$ for a task, an error on any $H_j$ implies an error for the entire set $\mathbf{H}$. To upper bound the probability of such a failure event, we consider the Bonferroni inequalities with odd low-order $k'<k$, which provides:
\begin{equation}
    \Pr(E) = \Pr\left(E_1 \cup ... \cup E_j\right) 
    \leq \sum_{r=1}^{k'} (-1)^{r+1} \sum_{(j_r) \in \mathcal{I}_r^{(k')}} \prod_{s=1}^r \Pr(E_{j_s})
\label{eq:bonferroni}
\end{equation}
where $\mathcal{I}_r^{(k')} := \{(j_r) \in \mathbb{N}^r : 1 \leq j_1 < \cdots < j_r \leq k' \}$.
This relation indicates that the probability of $\mathbf{H}$ having an error is bounded by a statistic of individual error probabilities, adjusted by correction terms that reflect their higher-order interactions. Following learning theory~\cite{zhang2023mathematical}, the probability of individual failures $E_j$ can be bounded in terms of the concept space size and sample count with $|\mathcal{H}|\exp(-2n\epsilon^2)$.
Let $\epsilon_j=err_{\mathcal{D}}(H_j)$ and $\epsilon'_j=\hat{err}_{\mathcal{S}_n}(H_j)$, we have:
\begin{equation}
    \Pr(E_j) = \Pr(\epsilon_j \geq \epsilon'_j + \epsilon) \leq |H_j| \exp\left\{-2n\epsilon^2\right\},
\label{eq:err_bound}
\end{equation}
where $n$ is the dataset size, and $|H_j|$ is the size of the concept space associated with $H_j$. 
Thus, to ensure that $\Pr(err_\mathcal{D}(\mathbf{H}) \leq \epsilon) \geq 1-\delta$, it is essential to control the individual error bounds $\Pr(E_j)$ and ensure sufficiently low overlap in their joint failure probability.

In the multi-modal case, data point $X=\{X_{m_\alpha},X_{m_\beta}\}\sim\mathcal{D}$ formed by the oracle pair $\{\mathcal{O}_\alpha$, $\mathcal{O}_\beta\}$, each emitting samples from modalities $m_\alpha$ and $m_\beta$, respectively. Both views share the same latent representation $K = \phi^*_\alpha(X_{m_\alpha}) = \phi^*_\beta(X_{m_\beta})$, enabling consistent labeling via shared step functions set $\textbf{H}$. This allows our theoretical framework and generalization bounds to extend naturally to multi-modal scenarios, as long as the encoder preserves semantic alignment across modalities.

% \newpage
\section{SCAR Data Quality Description}     
The SCAR scheme begins by estimating the FDS for each step function to ensure generalization with a controlled error. 
To characterize the minimal data size needed for generalization within a given error tolerance, we first introduce the definition of FDS:
\begin{definition}[Foundation Data Size]
    Given a dataset $D$ with distribution $\mathcal{D}$ and a target $\mathcal{H}$ with model $f$, $\forall \;\epsilon > 0$, $\exists\;n$ such that when $N > n$, the general error satisfies $\Pr(err_{\mathcal{D,H}}(f,D) \geq \epsilon)=0$, then $n$ is defined as the \emph{Foundation Data Size (FDS)}.
\label{def:foundation_data}
\end{definition}
In this section, we present a model-based approach to estimate this requirement.
We use a parametric model of concept growth and define SCAR measures.
These measures allow us to estimate the FDS under both independent and correlated error conditions.

\subsection{Estimation of Concept Space Size}
The tendency of model performance to saturate with data size motivates the following two assumptions on the growth of the concept space:
\textbf{First}, the size of the concept space increases monotonically with data size and eventually approaches an upper bound, which is commonly referred to as a \textit{scaling law} in some literature;
\textbf{Second}, each data increment independently contributes to reducing the gap between the current concept space and its upper bound.
Formally, we model the concept space size $|H_j|$ as $|H_j| = |H_j|^* \cdot R_{|H_j|}(n)$, where the rate function $R_{|H_j|}(n)$ satisfies the recursive property:
\begin{equation}
    1 - R_{|H_j|}(n_1 + n_2) = (1 - R_{|H_j|}(n_1)) \cdot (1 - R_{|H_j|}(n_2)),
\end{equation}
which yields the general solution: $R_{|H_j|}(n) = 1 - \alpha^{-\lambda_j n}$.
This formulation is motivated by the exponential dependence of the generalization bound on the data size, as shown in Eq.~\ref{eq:err_bound}.
Accordingly, the estimated concept space size can be rewritten as:
\begin{equation}
    |H_j| = \alpha_j (1 - e^{-\lambda_j n}),
\label{eq:concept_size}
\end{equation}
where $\alpha_j, \lambda_j > 0$ are regression parameters.
Building on the hypothesized growth behavior, we leverage upper bounds based on Eq.\ref{eq:err_bound} to guide the estimation of the concept space’s limiting value. To ensure that $\Pr(\epsilon_j \geq \epsilon'_j + \epsilon)\leq\delta$ holds, we can have:
\begin{equation}
  |H_j| \leq \delta \exp\left\{2n\epsilon^2\right\}.
\label{eq:concept_bound}
\end{equation}
This gives a theoretical upper bound on $|H_j|$ for fixed $n$, error gap $\epsilon$, and confidence level $\delta$, influenced by data quality during ERM optimization. By evaluating model performance on downsampled data, we apply Eq.~\ref{eq:err_bound} to constrain the growth curve and estimate the asymptotic concept space size $|H_j|^* = \alpha_j$ via regression.

\subsection{SCAR Measures Calculation}
Building on the assumptions and theoretical bounds, we quantify the influence on concept space estimation by examining performance trends under controlled downsampling. We propose the SCAR measures, \textbf{Scale}, \textbf{Coverage}, \textbf{Authenticity}, and \textbf{Richness}, each reflecting a distinct dimension of data characteristics within step-function modeling. To characterize a target dataset $\mathcal{S}_n^*$ of size $n^*$, we generate subsets $\mathcal{S}_n$ with size $n = n^* \cdot r$.

To assess data quality from a model-specific view, we use a fixed pre-trained encoder $\phi(\cdot)$ to extract embeddings $e = \phi(X)$ that approximate the underlying knowledge $K$.
For each subset $\mathcal{S}_n$, we define step function labels ${Y_k}$ via clustering or external labels, and evaluate them with a linear classification boundary\footnote{We use a single-layer MLP for efficiency.}.
SCAR measures are then computed across these step functions, with dataset partitions for each $H_j \in \mathbf{H}$ defined as in Eq.~\ref{eq:bonferroni}.
\begin{align*}
  X_{V_j} &= \{X_i \in \mathcal{S}_n \mid \forall j=1,...,k,\; \hat{H}_j(X_i) = Y_{ij} \}, \\
  X_{S_j} &= \{X_i \in \mathcal{S}_n \mid X_i \notin X_{V_j},\; \exists j=1,...,k,\; \hat{H}_j(X_i) = Y_{ij} \}, \\
  X_{I_j} &= \{X_i \in \mathcal{S}_n \mid \forall j=1,...,k,\; \hat{H}_j(X_i) \neq Y_{ij} \}.
\end{align*}
where $X_{V_j}$, $X_{S_j}$, and $X_{I_j}$ denote the valid, semi-valid, and invalid subsets, respectively.  
The SCAR indices are computed based on these partitions with eqs.~\ref{eq:concept_bound} and \ref{eq:concept_size}.

\subsubsection{\textbf{Scale}}
The data size directly affects the general bound, and Eq.\ref{eq:concept_size} implies that computing $n$ and $r$ yields numerically equivalent results with $\lambda$.
We use the down-sampling ratio $r$ to represent the scale of each subset in our analysis: $I_{\text{s}} \triangleq r$ and $n = r \cdot n^*$.

\subsubsection{\textbf{Coverage}}
Coverage quantifies the semantic diversity of data in the embedding space. 
Higher coverage indicates more stable decision boundaries.
We estimate this by computing the Jensen-Shannon Divergence between the distribution of logit values (within the same class) and a fitted Gaussian $\mathcal{N}(\mu, \sigma)$, where $\mu$ and $\sigma$ are the empirical mean and variance.
\begin{equation*}
  I_{\text{c}} = \frac{1}{2} \left(  
    \mathrm{KL}(D(\mathcal{S}_n) \,\|\, M) + \mathrm{KL}(\mathcal{N}(\mu, \sigma) \,\|\, M) 
  \right)
  \triangleq 1 - \delta,
\end{equation*}
where $M = \frac{1}{2}(D(\mathcal{S}_n) + \mathcal{N}(\mu, \sigma))$ is the mixture distribution between the empirical distribution $D(\mathcal{S}_n)$ and the Gaussian.  
The coverage informs the confidence $\delta$ associated with the error event $E_j$.

\subsubsection{\textbf{Authenticity}}
Authenticity measures the accuracy of the indicator function $H_j$ under the current embedding $\phi(\cdot)$:
\begin{equation*}
  I_{\text{a}} = \frac{1}{n}\sum_i \mathbb{I}(\hat{H}_j(X_i)=Y_{ij})
   \triangleq1 - \epsilon_j
\end{equation*}
where $\mathbb{I}(\cdot)$ is the indicator function.  
It reflects the strict classification error $\epsilon_j$ and serves as a key factor in estimating concept space size.

\subsubsection{\textbf{Richness}}
Richness complements strict accuracy by accounting for semi-valid samples—those misclassified by one indicator but correctly predicted by others.  
It captures useful semantic signals from partially correct predictions.  
Formally, richness is defined as:
\begin{equation*}
  I_{\text{r}} = \left(|X_{V_j}| + |X_{S_j}|\right) / |S_n|
  \triangleq 1 - \epsilon_j'
\end{equation*}
As a loose error measure, richness approximates the empirical loss used in estimating the concept space size.

\subsection{Estimation of Foundation Data}
Based on the SCAR measures, we here estimate the FDS required for a given signal function \(H_j\) in the concept space $\mathbf{H}$ with conditional regression. From Eq.~\ref{eq:concept_bound} and Eq.~\ref{eq:concept_size}, we have:
\begin{align}
    |H_j| &\leq I_{c}\exp\{2I_{s}(I_{r}-I_{a})^2\} \\
    |H_j| &= \alpha_j(1 - \exp\{-\lambda_j^* I_{s}\}),\;\;|H_j|^*=\alpha_j
\end{align}
where $\lambda_j^* = \lambda_j \times n$ is numerically equivalent to $\lambda_j$ in regression and does not affect the estimation of $|H_j|^*$, the concept space size of the signal function $H_j$.
Considering the Eq. \ref{eq:err_bound}, we can derive the following inequality:
\begin{equation*}
    n_j \geq \ln(|H_j|^*/\delta^*) \ 2\epsilon^{*2}
\end{equation*}
With the ideal set of parameters $\delta^*$ and $\epsilon^*$, We derive the lower bound on the sample size required to ensure a bounded generalization error, which serves as an estimate for the FDS for the single-step function\footnote{We approximate Def.~\ref{def:foundation_data} using $\delta = 0.01$ and $\epsilon = 0.01$ in practical settings.}.

% \newpage
\section{Data Characteristics Description}

\begin{figure}[t]
    \centering
    \includegraphics[width=0.475\textwidth]{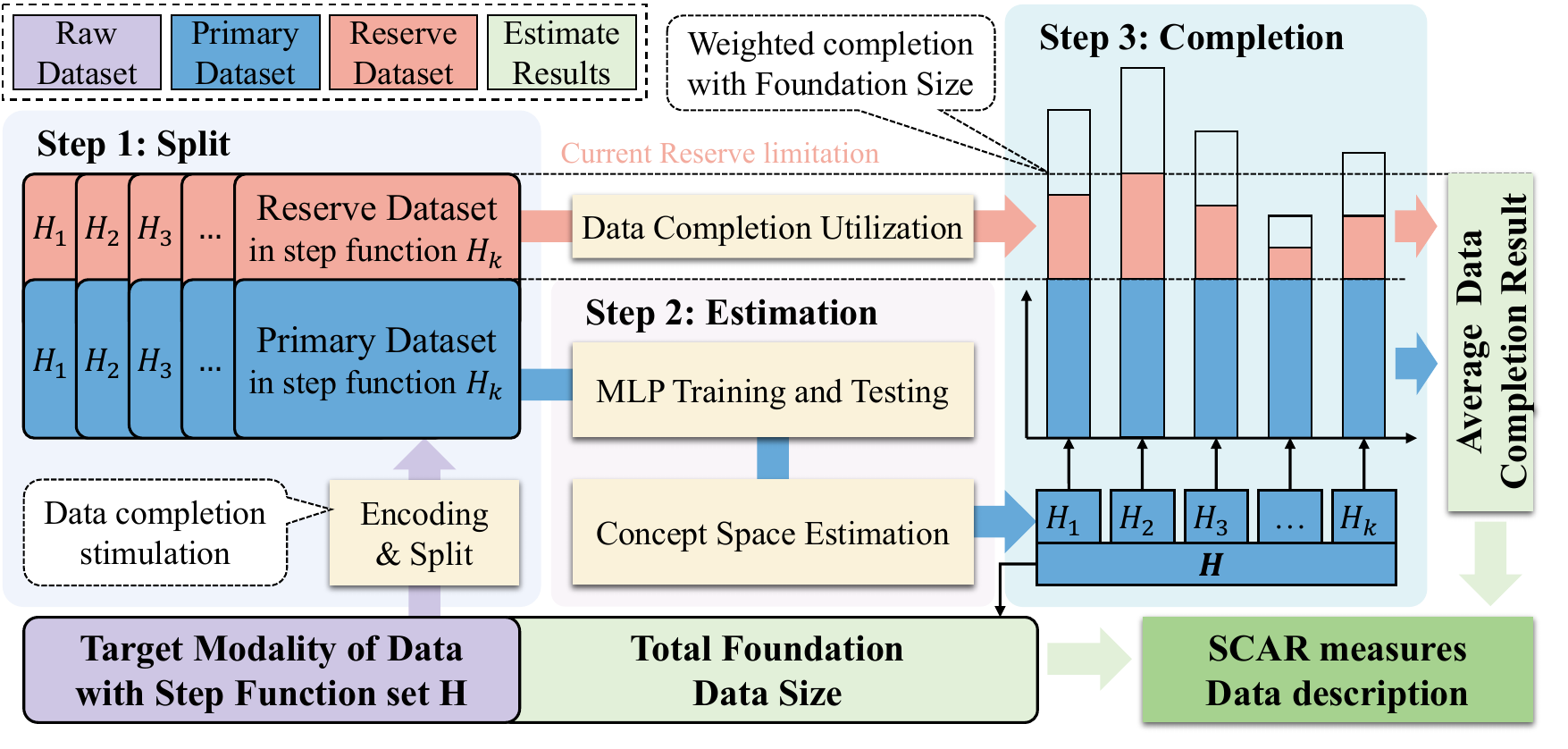}
    \caption{Overview of the SCAR-guided data completion and description process. The target dataset with step function set $H$ is split into primary and reserve subsets for concept space estimation and MLP-based evaluation. The estimated foundation sizes are used to compute a weighted data completion score across all step functions.}
    \Description{1234}
    \label{fig:process}
\end{figure}

In Section~3, we estimate the concept space and foundation size (FDS) for each step function $H_j$. We then extend this to the entire set $\mathbf{H}$ by modeling their joint structure, enabling set-level concept and foundation estimation. Based on this, we propose a SCAR-guided strategy for dataset expansion and define a unified SCAR measure to characterize the overall properties of the target modality. The SCAR process for a single modality is illustrated in Figure~\ref{fig:process}.

\subsection{FDS Estimation of Function Set}
Building on Eq.~\ref{eq:bonferroni}, which links individual step functions to the overall concept space, we derive a method to estimate the concept space size and FDS at the set level.
Prior work \cite{northcutt2021confident, jiang2021information} on learning with noisy labels has shown that label noise introduces systematic errors that lead to \emph{correlated failures} across step functions during the ERM process.  
As a result, the error events $\{E_j\}$ are \emph{positively correlated}, such that for any $k' < k$, the joint probability satisfies:
\begin{equation}  
    \Pr\left(E_1\cap...\cap E_j\right) \geq \prod\nolimits_{j=1}^{k'} \Pr(E_j),
    % \Pr\left( \bigcap_{j=1}^{k'} E_j \right) \geq \prod_{j=1}^{k'} \Pr(E_j),
\end{equation}
reflecting the dependency structure induced by noisy supervision.
To estimate the total failure probability $\Pr(E)$, we adopt the bound $\Pr(E_j)\leq |H_j^*| \exp(-2n\epsilon^2)$ from Eq.~\ref{eq:bonferroni}, and define $t = \exp(-2n\epsilon^2)$.  
Substituting into a Bonferroni-type inequality, we derive:
\begin{equation}
    \Pr(E) \leq \sum_{r=1}^{k'} (-1)^{r+1} \sum_{(j_r) \in \mathcal{I}_r^{(k')}} t^r \prod_{s=1}^r |H_{j_s}^*|
\label{eq:bon_multi}
\end{equation}
Assuming $\Pr(E) \leq \delta^*_E$ holds, we analyze the right side as a function of $t \in (0,1)$.
There exists a threshold $t^*$ at which the bound is attained with equality in step function series.
Solving $\exp(-2n^* \epsilon_E^{*2}) = t^*$ numerically for a given $\epsilon^*_E$ yields a lower bound $n^*$ on the FDS required to support the full set $\mathbf{H}$ under correlated noise\footnote{We use $r=3$, $\delta^*_E = 0.01$ and $\epsilon^*_E = 0.01\cdot\log(k)$ in our experiments.}.

\subsection{Data Completion Based on SCAR}
Based on the previously estimated FDS, we perform targeted data completion to meet task-specific requirements
For a dataset of size $n$, with $n_j$ samples aligned to function $H_j$, we estimate the required total FDS $n^*$ and the per-function requirement $n_j^*$.
To determine the number of additional samples needed for each $H_j$, we account for overlapping coverage across functions and distribute the total shortfall proportionally as follows:
\begin{equation*}
    n_j' = w_j\cdot \max(0, n^* - n),\;\;
    % w_j=\max(0, n_j^* - n_j)/\sum_j \max(0, n_j^* - n_j),
    w_j=\frac{\max(0, n_j^* - n_j)}{\sum_j \max(0, n_j^* - n_j)} 
\label{eq:scarless}
\end{equation*}
where $n_j'$ denotes the normalized allocated number of additional samples for $H_j$, normalized across all underrepresented functions.

\begin{algorithm}[t]
\caption{SCAR-Guided Foundation Estimation and Extension}
\begin{algorithmic}[1]
\State \textbf{Input:} Dataset $\mathcal{D}$, Encoder $E$, Split ratio $r$, sampling ratios $S$
\State \textbf{Output:} Foundation size $N_{F_T}$, SCAR indices $I_{scar}=\{S,C,A,R\}$

\State Initialize dataset classifier $M_D$ with encoder $E$

\State \textbf{$\#$ 1. Total Set Evaluation (Multi-Modality)}
\State Extract modality embeddings using encoder set $\{E_{\text{modal}}\}$
\For{each modality $m$}
    \State Run $k$-means on  $\mathcal{D}(m) \to y^{\text{m}}$, define step functions $H_j^{\text{m}}$
% \State Run $k$-means on text $\to y^{\text{text}}$, define step functions $H_j^{\text{img}}$

    \State Estimate $\{N_{F}^{\text{m}}\}$ using SCAR for $\mathcal{D}(m)$

    \State \textbf{$\#$ 2. Primary / Reserve Split}
    \State Split $\mathcal{D}_{\text{train}}(m) \to (\mathcal{D}_p(m), \mathcal{D}_r(m))$ using $r$
    \State Estimate each step foundation sizes $\{n_j^*\}$ from $\mathcal{D}_p(m)$
    \State Estimate the total foundation sizes $n^*$ from $\{n_j^*\}$

    \State \textbf{$\#$ 3. Per-Class Fill Size Estimation}
    \State Count required fill size: $\text{fill}_j(m) = \max(0, n^*_j - |\mathcal{D}_{p,j}(m)|)$
    \State Count $r_j=\text{fill}_j(m) / \sum_j \text{fill}_j(m)$
    \State Count available reserve size: $n_r = |\mathcal{D}_{r,j}(m)|/\max\{r_j\}$
    \For{each step index $j$}
        \State Select $n_r\cdot r_j$ samples from $\mathcal{D}_{r,j}(m)$ 
        \State $\mathcal{D}_{\text{ext},j}(m) \gets \mathcal{D}_{p,j}(m) \cup$ sampled data
    \EndFor
    \State \textbf{$\#$ 4. Final Evaluation}
    \State Run EMR process on $\mathcal{D}_{\text{ext}}(m) \to$ test accuracy $A_{\text{ext}}(m)$
    \State $S, C, A, R \gets N_T/N_{F_T}, N_T/N_{E}, A_{\text{ext}}(m), N_{F_T}/N_{T}$
\EndFor
\end{algorithmic}
\label{alg:scar}
\end{algorithm}

\subsection{General Data Characteristics Description}
While the SCAR indices were initially introduced to estimate the FDS for individual step functions, we here extend them to describe dataset characteristics from a broader perspective. Given a dataset $\mathcal{D}$ with $n$ samples and $k$ classes, we apply a class-aware partition with ratio $r$ to obtain a primary subset $\mathcal{D}_p$ and a reserve subset $\mathcal{D}_r$. A representation-driven completion process augments $\mathcal{D}_p$ using samples from $\mathcal{D}_r$, yielding an extended set $\mathcal{D}_e$. The $n^* = \min\{n_j'/w_j|n_j'\in\mathcal{D}_r,w_j>0\}$ is the bottleneck resource in the reserve subset, and the SCAR measures are defined as follows:
\begin{equation*}
    I_s = \frac{\min(n, n^*)}{n^*},\;
    I_c = \frac{n_e - n_p}{n - n_p},\;
    I_a = \text{Acc}_{\mathcal{D}},\;
    I_r = \frac{\min(n, n^*)}{n}.
\end{equation*}

Here, $I_s$ measures the closeness between the current dataset size and the estimated FDS $n^*$, reflecting whether the dataset meets the basic quantity requirements of the task. 
$I_c$ quantifies how effectively the reserve set complements the primary set during the completion process, capturing the dataset's representational coverage---i.e., the extent to which additional samples provide diverse and compatible information.
$I_a$ denotes model accuracy on the original dataset, reflecting current task fitness. 
Finally, $I_r$ evaluates the alignment between the dataset’s label granularity and its potential task capacity, indicating whether the current representation supports fine-grained task distinctions.
SCAR comprehensively characterizes a dataset’s essential properties by jointly evaluating its sufficiency, coverage, accuracy, and resolution with task-oriented generalization.

To extend this supplementation strategy to multimodal data, we adopt a cross-modal pseudo-supervision approach to estimate FDS in the shared image-text embedding space.  
We first apply $k$-means clustering to one modality (e.g., images) to obtain pseudo-labels $y_i^{\text{img}} \in \{1, \dots, k\}$, which define step function targets for the paired modality via image-to-text and text-to-image pseudo-supervision:
\begin{equation*}
    H_j^{\text{text}}(x_i^{\text{text}}) = \mathbb{I}[y_i^{\text{img}} = j],\quad H_j^{\text{img}}(x_i^{\text{img}}) = \mathbb{I}[y_i^{\text{text}} = j].
\end{equation*}
This bidirectional pseudo-supervision yields two FDS estimates, $n^*_{\text{img}}$ and $n^*_{\text{text}}$, for the image and text modalities.   
This approach enables SCAR to guide cross-modal data completion while respecting modality-specific concept distributions.
Algorithm~\ref{alg:scar} outlines the full SCAR-based data quality evaluation procedure.

% \newpage
\section{Experiments}

\subsection{Settings}

We evaluate the SCAR scheme with various data types, including text, image, video, and audio, under label or alignment supervision using selected pretrained models.
Tab.~\ref{tab:task_benchmark_encoder} summarizes the datasets, encoders, and training configurations used in our experiments.

\begin{table}[ht]
\centering
\caption{Summary of tasks, datasets, and pretrained encoders.}
\label{tab:task_benchmark_encoder}
\resizebox{0.475\textwidth}{!}{
\begin{tabular}{c|c|c|c|c}
\toprule
\hline
\textbf{Task} & \textbf{Encoders} & \textbf{Dataset} & \textbf{Train Size} & \textbf{\#Task Classes} \\
\hline
\multirow{3}{*}{Image} 
    & ResNet50\cite{resnet} & CIFAR-10          & 50k    & 10 \\
    & ViT-B/16\cite{vit}    & CIFAR-100         & 50k    & 100 \\
    & DINO-v2\cite{dino}    & ImageNet-1K       & 1.2M   & 1000 \\
\hline
\multirow{3}{*}{Text} 
    & BERT\cite{bert}       & AG-News           & 120k   & 4 \\
    & RoBERTa\cite{RoBERTa} & DBPedia Ontology  & 560k   & 14 \\
    & GPT2\cite{gpt}        & Wikipedia Subset  & 945k   & 100 \\
\hline
\multirow{3}{*}{Image-Text} 
    & CLIP\cite{clip}           & \multirow{3}{*}{\makecell[c]{Flickr30k \\ COCO Captions}} 
                                & \multirow{3}{*}{\makecell[c]{24k \\ 82k}} 
                                & \multirow{3}{*}{\makecell[c]{10 \\ 50}}   \\
    & CoCa\cite{coca}           & & & \\
    & SigLIP\cite{siglip}       & & & \\
\hline
\multirow{2}{*}{Video-Text} 
    & VideoCLIP\cite{videoclip} & \multirow{2}{*}{MSR-VTT}    & \multirow{2}{*}{9k}   & \multirow{2}{*}{3} \\
    & X-CLIP\cite{xclip}        &                             &                       &                    \\
\hline
\multirow{2}{*}{Audio-Text} 
    & CLAP\cite{clap}           & \multirow{2}{*}{AudioCaps}  & \multirow{2}{*}{45k}  & \multirow{2}{*}{5} \\
    & Pengi\cite{pengi}         &                             &                       &                    \\
\hline
\bottomrule
\end{tabular}}
\end{table}

Given the extracted embeddings from each dataset and encoder, we first estimate the total foundation size $N^*_T$ to estimate the ideal data scale of each dataset-model pair. To simulate data supplementation, we split the dataset into a 60\% primary set and a 40\% reserve set. SCAR measures are computed on the primary set to estimate the per-step foundation size $N_j^*$ via a linear model and SCAR measures for a single-step function. Based on normalized completion ratios, we select data from the reserve set to extend the data and assess the performance improvement.

We compare two data completion baselines: random sampling and class-wise averaging. All models use a single-layer linear classifier trained with softmax and cross-entropy loss, optimized by Adam (lr=1e-3, batch size=500) with early stopping (patience=5, max epochs=100). Results are averaged over three runs with mean and standard deviation reported.

\subsection{Experiment Results}

\begin{figure*}[t]
    \centering
    \begin{subfigure}[t]{0.54\textwidth}
        \centering
        \includegraphics[width=\textwidth]{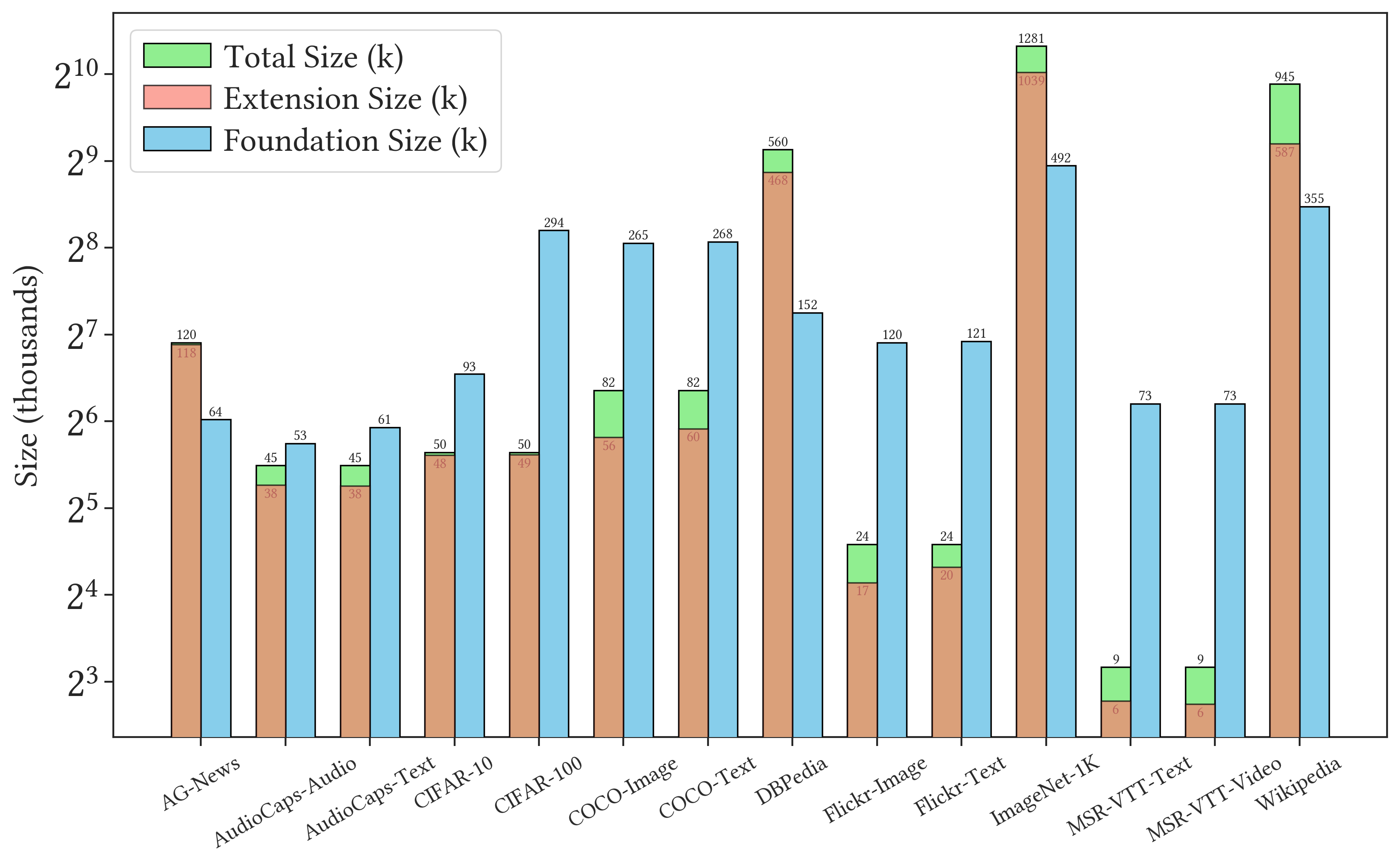}
        \caption{Data size for each Dataset-Model pair.}
        \Description{1234}
        \label{fig:scar-size}
    \end{subfigure}
    \begin{subfigure}[t]{0.44\textwidth}
        \centering
        \includegraphics[width=\textwidth]{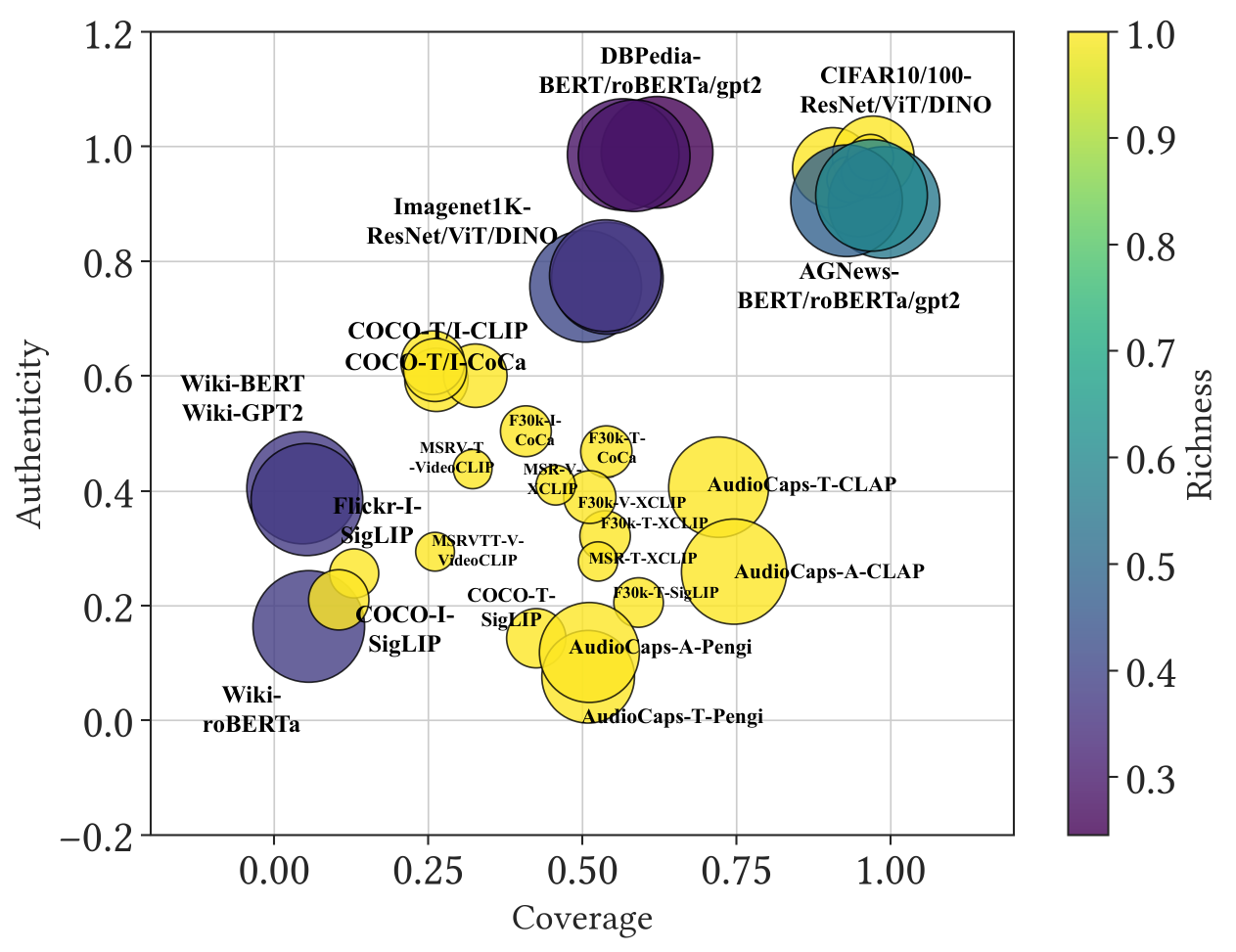}
        \caption{SCAR-based quality metrics.}
        \Description{1234}
        \label{fig:scar-bubble}
    \end{subfigure}
    \caption{Comparison of dataset-model pairs based on SCAR: (a) Data scale varies across pairs, with foundation size dominating in large datasets. (b) SCAR metrics reveal modality-dependent trade-offs among scale, coverage, richness, and authenticity.}
    \label{fig:scar-overall}
\end{figure*}

Figure~\ref{fig:scar-size} presents the average SCAR-estimated data sizes across all models for each dataset-model pair, including the Total (orange), Extension (green), and Foundation (blue) components. In general, large-scale datasets such as ImageNet-1K, Wikipedia, and the DBpedia ontology typically meet the size requirements for Foundation Data. However, the observed gap in data completion suggests non-uniform model fitting across different classes within these datasets. In contrast, popular datasets like CIFAR-10/100, AG-News, and AudioCaps exhibit minimal differences between their total and extended sizes, indicating better balance in class-level representation. COCO Caption shows a larger imbalance in this regard. Smaller datasets such as Flickr30k and MSR-VTT fall significantly short of the Foundation Data size and also display greater representational bias compared to others. These results highlight the uneven representational balance across datasets, stressing the need for modality- and scale-aware data completion.

Figure~\ref{fig:scar-bubble} further reveals distinct modality- and domain-specific patterns through SCAR metrics, including scale (bubble size), coverage, authenticity, and richness (color-coded). Notably, large-scale datasets like ImageNet-1K and DBPedia exhibit high authenticity but differ in coverage and richness, reflecting domain-specific sampling biases. In contrast, well-balanced datasets such as CIFAR10/100 and AG-News show consistent coverage and moderate scale, suggesting stronger representational balance. Meanwhile, multimodal datasets such as AudioCaps and COCO exhibit significant variance in both authenticity and coverage across their text and image components, indicating uneven cross-modal alignment. These observations confirm that SCAR effectively captures the nuanced characteristics of dataset-model pairs from multiple perspectives, offering a principled basis for evaluating and guiding data acquisition and completion strategies.

\begin{table}[t]
\centering
\caption{Results of SCAR-based data completion experiment of multi-modal data description.}
\label{exp:main}
\resizebox{0.47\textwidth}{!}{
\begin{tabular}{ccc|ccc}
\toprule
\hline
\multirow{2}{*}{Dataset} & \multirow{2}{*}{Encoder} & \multirow{2}{*}{Pri. Acc.~($\uparrow$)} & \multicolumn{3}{c}{Ext. Acc.~($\uparrow$)} \\
\cline{4-6}
                         &                          &                                         & SCAR & rand. & avg. \\
\hline
\multirow{3}{*}{\makecell{Flickr30k (Text)}}
    &        CLIP\cite{clip}         & $24.66_{\pm8.34}$ & $\bm{41.65_{\pm23.20}}$ & $30.72_{\pm18.69}$ & $38.14_{\pm17.85}$ \\
    &        CoCa\cite{coca}         & $38.30_{\pm15.67}$ & $\bm{57.86_{\pm0.92}}$ & $37.31_{\pm30.23}$ & $48.04_{\pm18.67}$ \\
    &        SigLIP\cite{siglip}     & $21.46_{\pm3.93}$ & $\bm{21.95_{\pm5.77}}$ & $21.89_{\pm6.55}$ & $23.92_{\pm14.52}$ \\
\cline{2-6}

\multirow{3}{*}{\makecell{Flickr30k (Image)}}
    &        CLIP\cite{clip}         & $40.26_{\pm1.09}$ & $\bm{45.97_{\pm0.69}}$ & $40.83_{\pm0.95}$ & $43.32_{\pm0.71}$ \\
    &        CoCa\cite{coca}         & $50.05_{\pm0.16}$ & $\bm{51.54_{\pm0.37}}$ & $49.97_{\pm0.45}$ & $51.03_{\pm0.53}$ \\
    &        SigLIP\cite{siglip}     & $29.60_{\pm0.18}$ & $\bm{29.69_{\pm0.06}}$ & $29.67_{\pm0.10}$ & $25.71_{\pm11.88}$ \\
\hline

\multirow{3}{*}{\makecell{COCO Captions\\(Text)}}
    &        CLIP\cite{clip}         & $59.62_{\pm0.39}$ & $\bm{62.27_{\pm0.98}}$ & $58.93_{\pm0.58}$ & $61.79_{\pm0.56}$ \\
    &        CoCa\cite{coca}         & $60.11_{\pm1.86}$ & $\bm{64.88_{\pm0.87}}$ & $59.79_{\pm1.96}$ & $64.19_{\pm1.40}$ \\
    &        SigLIP\cite{siglip}     & $14.43_{\pm3.60}$ & $\bm{18.37_{\pm0.39}}$ & $13.69_{\pm3.25}$ & $15.83_{\pm4.78}$ \\
\cline{2-6}

\multirow{3}{*}{\makecell{COCO Captions\\(Image)}}
    &        CLIP\cite{clip}         & $62.09_{\pm0.68}$ & $\bm{66.03_{\pm1.17}}$ & $62.76_{\pm0.37}$ & $65.75_{\pm0.74}$ \\
    &        CoCa\cite{coca}         & $60.81_{\pm1.20}$ & $62.64_{\pm0.32}$ & $61.19_{\pm0.19}$ & $\bm{62.72_{\pm0.42}}$ \\
    &        SigLIP\cite{siglip}     & $21.24_{\pm0.53}$ & $\bm{21.32_{\pm1.15}}$ & $20.07_{\pm3.28}$ & $21.31_{\pm5.98}$ \\
\hline

\multirow{2}{*}{\makecell{MSR-VTT (Text)}}
    & VideoCLIP\cite{videoclip} & $33.33_{\pm31.30}$ & $42.03_{\pm5.20}$ & $42.03_{\pm5.20}$ & $42.03_{\pm5.20}$ \\
    & X-CLIP\cite{xclip}        & $44.97_{\pm1.00}$ & $\bm{44.17_{\pm3.40}}$ & $33.97_{\pm17.00}$ & $36.80_{\pm14.70}$ \\
\cline{2-6}

\multirow{2}{*}{\makecell{MSR-VTT (Video)}}
    & VideoCLIP\cite{videoclip} & $25.90_{\pm3.50}$ & $\bm{31.27_{\pm7.90}}$ & $29.10_{\pm10.00}$ & $29.37_{\pm4.30}$ \\ 
    & X-CLIP\cite{xclip}        & $43.67_{\pm7.30}$ & $41.30_{\pm5.60}$ & $\bm{43.10_{\pm4.20}}$ & $38.60_{\pm3.50}$ \\
\hline

\multirow{2}{*}{\makecell{AudioCaps (Text)}}
    & CLAP\cite{clap}           & $41.06_{\pm0.27}$ & $40.47_{\pm0.36}$ & $\bm{40.97_{\pm0.81}}$ & $40.94_{\pm0.18}$ \\
    & Pengi\cite{pengi}         & $7.72_{\pm0.18}$ & $7.24_{\pm0.14}$ & $\bm{7.69_{\pm0.23}}$ & $7.48_{\pm0.40}$ \\
\cline{2-6}

\multirow{2}{*}{\makecell{AudioCaps (Audio)}}
    & CLAP\cite{clap}           & $26.56_{\pm0.54}$ & $25.99_{\pm1.03}$ & $\bm{26.27_{\pm0.90}}$ & $25.91_{\pm0.22}$ \\
    & Pengi\cite{pengi}         & $12.07_{\pm1.84}$ & $11.04_{\pm0.05}$ & $\bm{11.91_{\pm1.58}}$ & $10.83_{\pm0.45}$ \\
\hline
\bottomrule
\end{tabular}}
\end{table}

The experimental results in Table~\ref{exp:main} demonstrate the effectiveness of SCAR-based data completion with optimal linear model testing results across multiple modalities and datasets. Compared to random and average-based baselines, SCAR consistently delivers superior or comparable performance improvements across vision, text, and audio domains. For instance, in datasets such as Flickr30k and COCO Captions, SCAR-enhanced extensions lead to noticeable accuracy gains, particularly for models like CLIP and CoCa. While improvements are less pronounced for models with limited initial performance or inherently balanced data (e.g., SigLIP or AudioCaps), SCAR still maintains parity or avoids performance degradation. These results validate the robustness of SCAR-guided data selection in identifying high-utility samples under diverse settings. Overall, SCAR provides a principled approach for augmenting datasets, particularly when additional data can meaningfully improve model performance, and avoids unnecessary expansion when no substantial gain is achievable.

% \newpage
\section{Conclusion}
This work represents an advancement toward principled data-centric learning for foundation models. 
By bridging learning theory with practical data quality modeling, we introduce SCAR—a structured framework that quantifies dataset utility within the concept space. 
The notion of Foundation Data and its estimation provides a theory-driven approach for adaptive dataset construction, moving beyond heuristic data collection and pruning strategies. 
Through extensive experiments spanning multiple modalities, SCAR effectively captures key dataset characteristics under scaling from various perspectives. 
This scheme lays the foundation for more generalizable foundation model pipelines, while fostering the development of scalable, high-quality, and generalization-focused data systems.

% \newpage
\bibliographystyle{acm}
\bibliography{reference}

\end{document}